\documentclass[conference]{IEEEtran}
\IEEEoverridecommandlockouts
\usepackage{cite}
\usepackage{amsmath,amssymb,amsfonts}
\usepackage{multicol,caption}
\usepackage{commath}
\usepackage{mathtools}
\usepackage{algorithmic}
\usepackage{graphicx}
\usepackage{textcomp}
\usepackage{hyperref}
\usepackage{multirow}
\usepackage{comment}
\usepackage{xcolor}
\usepackage[
  separate-uncertainty = true,
  multi-part-units = repeat
]{siunitx}
\def\BibTeX{{\rm B\kern-.05em{\sc i\kern-.025em b}\kern-.08em
    T\kern-.1667em\lower.7ex\hbox{E}\kern-.125emX}}

\usepackage[utf8]{inputenc}
\usepackage[english]{babel}
\usepackage{authblk}
\newtheorem{theorem}{Axiom} [section]

\begin{document}
\title{Spatially Correlated Patterns in Adversarial Images}
\author{Nandish Chattopadhyay} \author{Lionell Yip En Zhi} \author{Bryan Tan Bing Xing} \author{Anupam Chattopadhyay}
\affil{School of Computer Science and Engineering, Nanyang Technological University, Singapore}

\maketitle

\begin{abstract}
Adversarial attacks have proved to be the major impediment in the progress on research towards reliable machine learning solutions. Carefully crafted perturbations, imperceptible
to human vision, can be added to images to force misclassification by an otherwise high performing neural network. To have a better understanding of the key contributors of such structured attacks, we searched for and studied spatially co-located patterns in the distribution of pixels in the input space. In this paper, we propose a framework for segregating and isolating regions within an input image which are particularly critical towards either classification (during inference), or adversarial vulnerability or both. We assert that during inference, the trained model looks at a specific region in the image, which we call Region of Importance (RoI); and the attacker looks at a region to alter/modify, which we call Region of Attack (RoA). The success of this approach could also be used to design a post-hoc adversarial defence method, as illustrated by our observations. This uses the notion of blocking out (we call neutralizing) that region of the image which is highly vulnerable to adversarial attacks but is not important for the task of classification. We establish the theoretical setup for formalising the process of segregation, isolation and neutralization and substantiate it through empirical analysis on standard benchmarking datasets. The findings strongly indicate that mapping features into the input space preserves the significant patterns typically observed in the feature-space while adding major interpretability and therefore simplifies potential defensive mechanisms.

\end{abstract}


\section{Introduction} \label{sec:introduction}
The growth in interest in machine learning and the data driven approach has advanced by leaps and bounces in recent times. This has benefited from the developments in research in the all-encompassing AI domain, both in industry and the academic institutions. A major push in the direction came from the breakthrough of neural networks achieving super-human performances in certain specific tasks in 2012 ~\cite{hinton}. Historically, traditional statistical learning methods were known to involve a lot of effort in feature-engineering and pre-processing data before modeling ~\cite{esl}. The performance in terms of prediction accuracy started to reach a saturated level with that approach and lacked generalization. 
The success of deep learning algorithms rejuvenated the community in paying greater attention to neural network architectures. Originally introduced in the '50s ~\cite{rosenblatt}, these learning mechanisms were not living up to their full potential due to severe lack in processing capabilities of the hardware resources. With the developments in multi-core hardware architectures and GPUs coming into the scene, training neural networks were much more feasible. The massive matrix-operations could easily be carried out using sequences of linear transformations which could run parallelly ~\cite{bishop}. Thereby, the enormous learning capacity of these networks could be harnessed.


\subsection{Background} \label{background}
The flourishing results in AI research however did not translate into reliable solutions as a key vulnerability of such models was observed soon after. In 2015, adversarial attacks on highly accurate neural network models were demonstrated ~\cite{szegedy}. A trained model would achieve very high accuracy on a clean test set but the performance would be very poor on a set of adversarial samples. These samples could easily be constructed using highly structured perturbations which would remain imperceptible to the human vision ~\cite{basic}. Additionally, adversarial samples generated with respect to one neural network architecture would also serve as an adversarial sample ~\cite{papernot} for a different machine learning classifier like SVM ~\cite{svm}. Therefore, a plethora of research was carried out developing mechanisms of attack subsequently ~\cite{FSGM, cw}.


This resulted in a cat-and-mouse chase between adversarial attacks and defenses, as both kept getting better ~\cite{sir_survey}. Taking a step aside, some enthusiasts have been interested in looking at the root cause of the problem from first principles and understand why these attacks take place rather easily. Goodfellow et al.~\cite{linear} argued that the inherent linearity in the neural networks could facilitate the generation of adversarial samples. Others point out a different rationale for it, ~\cite{dube}, primarily addressing the curse of dimensionality ~\cite{googledimension, chattopadhyay}.


\subsection{Motivation} \label{Motivation}
Image classifiers, particularly neural networks optimize on a very complex landscape, wherein a decision boundary is obtained that separates the trained manifolds. The behaviour and properties of the classifier is closely related to the nature of the feature space ~\cite{TPAMI}. Adversarial attacks introduce small (bounded) targeted perturbations to the images so that the adversarial samples cross over the decision boundary, and is therefore misclassified by the model. The feature space is high dimensional and difficult to interpret. Therefore, it is necessary to map the properties and knowledge about the feature space on to the input space, so that the patterns are less abstract and can be used easily for various purposes, including the critical task of being robust against adversarial attacks. It is important to study if the properties and patterns of the feature space are preserved in the input space. 

\subsection{Contribution} \label{Contribution}
While analysing patterns of features for adversarial vulnerability, a pair of `Source' and `Target' classes is usually considered, since it is conceived that there is a specific path of movement of the samples in the high dimensional space across the decision boundary. Also, this is consistent with all samples belonging to the particular pair of classes. We take cognizance of these facts in our analysis. In this paper, we have formalised theoretically and thereafter verified empirically, a comprehensive study on:

\subsubsection{Segregation}

Use of spatially correlated patterns to identify and segregate regions within the input space (consisting of pixels of the image) specifically contributing towards:
    \begin{itemize}
        \item the correct classification of the input image during inference by the trained classifier (we call this region the Region of Importance (RoI))
        \item incorrect classification of the adversarial sample (generated through some adversarial attack) by the trained classifier (we call this region the Region of Attack (RoI))
    \end{itemize}
    
\subsubsection{Isolation}
Use this set of regions (RoI and RoA) to isolate four disjoint sets of pixels (we call $UV, U\bar{V},  \bar{U}V , \bar{U}\bar{V} $ based on their impact on utility $U$ and vulnerability $V$) which help us in explaining the reason as to why certain segments of the image are particularly more vulnerable to any adversarial attack than the rest.

\subsubsection{Neutralization}
Extract the region of most vulnerability from the thus obtained segments and attempt to block out its information content to establish a post-hoc adversarial defence mechanism. Specifically, we study the impact of the \textit{neutralization} process on:
    \begin{itemize}
        \item the image's classification by the trained classifier upon being modified
        \item the image's adversarial vulnerability upon being modified
    \end{itemize}

We have substantiated the propositions with thorough experimental verification on multiple standard datasets using a well-known classifier and attack mechanism.

\subsection{Organization} \label{Organization}
In Section \ref{theory}, we briefly touch upon some related work in the area of studying correlated features and its role in adversarial attacks and present the necessary theoretical framework, mentioning the assumptions and axioms underlying our proposition. In Section \ref{patterns}, we specify the details of the patterns that we intend to study in order to understand spatial co-location of pixels contributing to successful adversarial attacks and how they can be manipulated for a post-hoc defense mechanism. In Section \ref{experiments}, we present our experimental results and their significance. We conclude with some key remarks in Section \ref{conclusion}.

\section{Theoretical Formulation} \label{theory}
Although the maximum advancements in the realm of deep learning has been in the domain of supervised image classification tasks ~\cite{deep}, this is also where the initial adversarial attacks were seen. Images have been highly vulnerable to being sensitive towards little strategic perturbations.



An adversarial attack is a mechanism used by a malicious entity to force erroneous decisions by the classifier, by modifying test samples carefully, whilst keeping the modifications imperceptible to the annotator.

\subsection{Adversarial Attacks} \label{adversarial_attack}

\begin{figure}[!ht]
\centering
\includegraphics[width=0.3\textwidth]{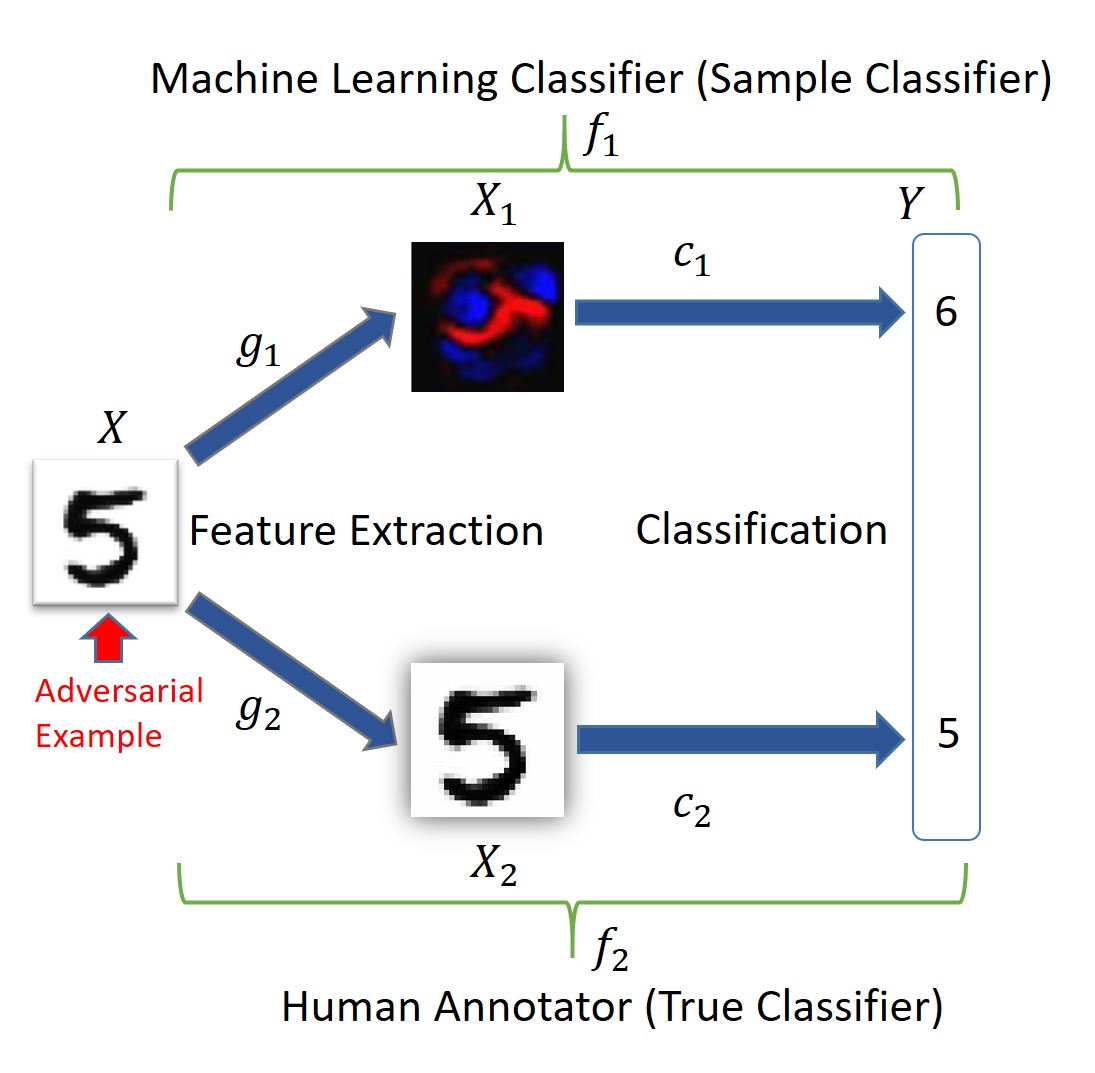}
\caption{Adversarial attacks on images}
\label{adv1}
\end{figure}
Let us consider $X$ to be the input sample space. 
There are two competing classifiers, $f_1$ (sample classifier) and $f_2$ (human annotator). The classifiers consist of two sections, for feature extraction and classification. We consider $X_1$ to be the feature space for the sample classifier and $X_2$ to be the feature space for the human annotator, where $d_1$ and $d_2$ are norms defined in the spaces $X_1$ and $X_2$ respectively. Therefore, $f_1=c_1 \circ g_1$ and $f_2 = c_2 \circ g_2$ as shown in Figure \ref{adv1}. 
If $x \in X$ is a randomly chosen training sample, then its corresponding adversarial example $x^*$, for a norm $d_2$ defined on the space $X_2$, and a predefined threshold $\delta > 0$, satisfies ~\cite{chattopadhyay}:
\begin{equation} 
\centering
\label{eq1}
\begin{split}
 f_1(x) \neq f_1(x^*) \quad \mbox{and} \quad f_2(x) = f_2(x^*)\\
 \mbox{such that} \quad d_2(g_2(x),g_2(x^*))< \delta \\  \nonumber 
\end{split}
\end{equation}

\subsection{Intuition for studying correlation among samples of a class} \label{corr}
The distribution of the data is important while studying adversarial attacks. A classifier, typically is a hyperplane that splits the data into trained manifolds, each pertaining to a particular class. The properties of the distribution of the data is easy to study within each of the manifolds. An adversarial example is a sample, that is made to shift across the hyperplane classifier by introducing structured perturbation. The geometry of the landscape of the features is therefore critical in the design of adversarial samples. For a particular sample belonging to its true class' manifold, in order to slide into another class' manifold thereby becoming an adversarial sample, it takes a specific path that is governed by the geometry of the two manifolds. This path will naturally be different for any pair of `Source' and `Target' classes, where a sample from a `Source' class is converted into an adversarial sample being misclassified into the `Target' class. The purpose of studying correlation among features is to look for a basis to observe patterns that can be associated with specific tasks like classification or adversarial attacks and the distribution of the samples in the corresponding optimization landscape.


\subsection{Significance of Features} \label{features}
Let us consider we have a binary classification setup, wherein we sample input-label pairs $(x,y) \in X \times \{\pm 1\}$ from a particular distribution $D$, for the purpose of training a classifier $C:X \to \{\pm1\}$, which is able to predict $y$, given an $x$. 

In an attempt to understand how features play a key role in classification and the generation of adversarial examples, let us define a \textit{feature} to be a map from the input space $X$ to $R$, therefore constituting a set of features $F=\{f:X \to R \}$.
The features can thereafter be categorised as proposed in the relevant literature ~\cite{features}. 

\textit{$\rho$-useful features}: Given a distribution $D$, a particular feature $f$ is said to be $\rho$-useful if the feature is correlated with the label of the class in expectation. Therefore, we have $E_{(x,y) \in D}[y \cdot f(x)] \geq \rho$. This follows that $\rho_{D}(f)$ is the largest $\rho$ for which the feature $f$ is $\rho$-useful given distribution $D$.  

\textit{$\gamma$-robustly useful features}: A $\rho$-useful feature $f(\rho_{D}(f) > 0$ is said to be \textit{robust} ($\gamma$-robustly useful feature for $\gamma > 0$) if when an adversarial perturbation $\Delta$ is introduced, $f$ remains $\gamma$-useful. That is, $E_{(x,y) \in D}[\inf_{\delta \in \Delta(x)} y \cdot f(x+\delta)] \geq \gamma$. 

\textit{Useful, non-robust features}: This is a $\rho$-useful feature which is not a $\gamma$-robust feature for any $\gamma \geq 0$. They make a contribution towards classification of the image, but have the potential to adversely affect accuracy in the adversarial setting.

\subsection{Interpretability: Feature space to Input space}

The existing work suggests correlation among features that play an important part in the process of both classification and the generation of adversarial examples. In this work, we study the relevance of that in the input space, instead of the feature space. The primary motivation behind that is the fact that the input space, comprising of pixels for an image, is more interpret-able and less abstract than the feature space or the high dimensional optimization landscape. For that purpose, we map the features into the input space and study how the correlation of features result into the co-location of pixels.

In order to study the mapping of the features of interest from the feature space to the pixels in the input space, we need to develop a framework. We can perceive the setup as a contest between two players, a classifier that tries to correctly classify a particular sample and an attacker that tries to mis-classify the same. 

Let us consider the input image as a set of pixels, each pixel is denoted by its location, indexed serially. Thus, let $X$ be the input image of $n$ pixels such that $X=\{x_{1}, \ldots, x_{n}\}$. The classifier maps $X$ to the set of labels $Y=\{\hat{y}, \Bar{y}\}$, such that for a particular `Source-Target' pair, $\hat{y}$ is the correct label and $\Bar{y}$ is the mis-classified label of the adversarial sample. The classifier looks at some pixels of the image contained in the set $X_{p}=x_{i}, \ldots, x_{p}$ for classification and the adversarial attacker modifies some pixels $X_{q}=x_{j}, \ldots, x_{q}$. Therefore, we have:
\begin{equation}
\begin{split}
    X_{p}= \{x_{i}, \ldots, x_{p} | x_{i}, \ldots, x_{p} \sim \hat{y} \} \\ \nonumber
    X_{q}= \{x_{j}, \ldots, x_{q} | x_{j}, \ldots, x_{q} \sim \Bar{y} \}
\end{split}
\end{equation}
where $\sim$ denotes that those particular pixels have leverage on the decision of the output. It must be noted here that $X_{p}$ and $X_{q}$ are not necessarily mutually exclusive. Based on this formulation, we can define two axioms. 
\begin{theorem} \label{axiom1}
For all input pixels $x_{k} \in X$, we assign $x_{k}$ to the Region of Importance ($RoI$) if $corr(x_{k}, \hat{y}) \geq \delta_{1}$, where $\delta_{1}$ is the threshold for the leverage of the pixel on the output label. That is, 
\begin{equation}
    \forall x_{k} \in X, \quad x_{k} \Rightarrow RoI \quad | \quad corr(x_{k}, \hat{y}) \geq \delta_{1} \nonumber
\end{equation}

Hence, we have $RoI \subseteq X_{p}$. 
\end{theorem}

\begin{theorem} \label{axiom2}
For all input pixels $x_{k} \in X$, we assign $x_{k}$ to the Region of Attack ($RoA$) if $corr(x_{k}, \Bar{y}) \geq \delta_{2}$, where $\delta_{2}$ is the threshold for the leverage of the pixel on the adversarial mis-classfied label. That is, 
\begin{equation}
    \forall x_{k} \in X, \quad x_{k} \Rightarrow RoA \quad | \quad corr(x_{k}, \Bar{y}) \geq \delta_{2} \nonumber
\end{equation}
Hence, we have $RoA \subseteq X_{q}$. 
\end{theorem}

Exactly like $X_{p}$ and $X_{q}$, the $RoI$ and $RoA$ are not mutually exclusive either. In fact, their overlap is of importance, as illustrated in the later sections of this paper.

Based on these axioms, we make the following assumptions theoretically:
\begin{itemize}
    \item The distribution of pixels belonging to RoI and RoA within an image is not random, the pixels are  correlated spatially, that is they are co-located. 
    \item The distribution of pixels belonging to RoI and RoA for images belonging to any particular class of images is not random, the pixels are correlated spatially. This is considered naturally for images which are translationally invariant because otherwise looking for spatial correlation across images in not very meaningful. 
\end{itemize}

We have tested for evidence regarding the validity of these two assumptions empirically on multiple datasets in Section \ref{experiments}. Having obtained the RoI and the RoA, we shall now proceed to further segmentation of the image into mutually exclusive regions, each with their own significance. It may also be noted here that, this approach preserves the ability of segregation of features/pixels into the binary categories of Usefulness and Robustness as well. We slightly modify the classification into `Utility' and 'Vulnerability', as shown in Figure \ref{split}.

\subsection{Segregating and Isolating Regions}
To accomplish the task of generating the RoI and the RoA (the process we call \textit{Segregation}) using the axioms \ref{axiom1} and \ref{axiom2}, we need a mechanism for studying the leverage of a particular pixel on the classification task and the adversarial attack. This would help us calculate the correlation and apply the thresholds. Considering a convolutional neural network based classifier (as it is most widely used for image classification tasks, although any neural network architecture is applicable here)  and any generic adversarial attack mechanism, we:
\begin{itemize}
    \item map the features from the final flattened layer of the CNN to the input space (using class activation maps), note its significance to generate $X_{p}$ and use the threshold  $\delta_1$ to include a particular pixel to the RoI.
    \item look at the pixels in the input space that have been modified to generate an adversarial sample to generate $X_{q}$, and include those in the RoA which have been altered (by $L_{2}$ norm) beyond the threshold $\delta_2$.
\end{itemize}

\begin{figure}[!ht]
\centering
\includegraphics[width=0.485\textwidth]{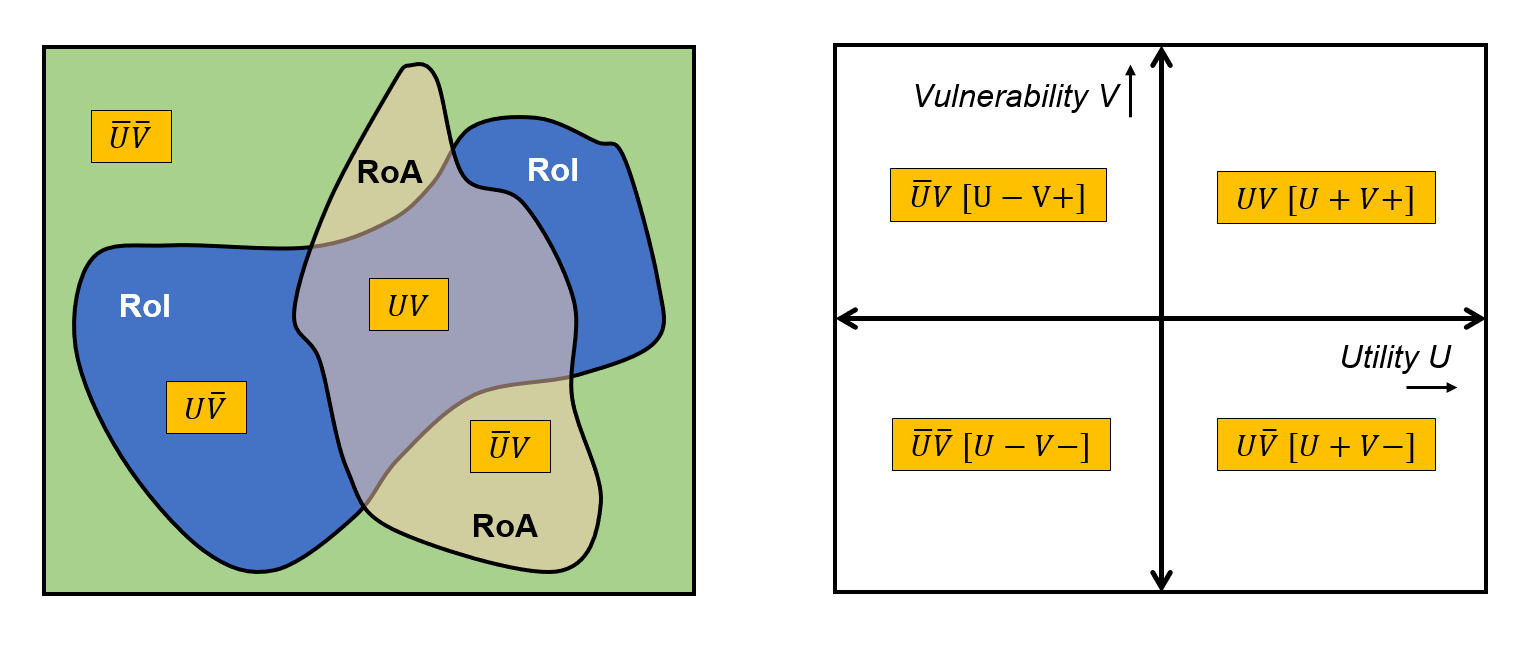}
\caption{Segmentation of the input pixels into regions, mapped as per their leverage on Utility $U$, Vulnerability $V$.}
\label{split}
\end{figure}

For any set of $(\delta_1,\delta_2)$, upon the generation of the RoI and the RoA, we can split the input space into four particular regions $UV, U\bar{V},  \bar{U}V , \bar{U}\bar{V} $  as shown hereafter, each with specific significance (based on utility U and vulnerability V, U+V+, U+V-, U-V+ and U-V- respectively). We call it \textit{Isolation}. It may be noted here that this \textit{Segregation} works for both images at the individual level with individual RoI and RoI, and also at the class level, using the representative ones, if the assumptions made above hold good. For the image $X$ containing $n$ pixels, each denoted by its corresponding position index number, we can construct the disjoint partitions $UV, U\bar{V},  \bar{U}V , \bar{U}\bar{V} $ as:
\begin{equation}
\begin{aligned}
    UV \quad [ U+V+ ] = \{x | x\in RoI, x\in RoA\} \\
    U\bar{V} \quad [U+V-] = \{x | x\in RoI, x\notin RoA\} \\
    \bar{U}V \quad [U-V+] = \{x | x\notin RoI, x\in RoA\} \\
    \bar{U}\bar{V} \quad [U-V-] = \{x | x\notin RoI, x\notin RoA\}
\end{aligned}
\end{equation}
Intuitively, each region in a particular image and its significance can be expressed as follows: Region $U\bar{V}$ [U+V-] consists of those spatially co-located pixels that have an important role to play in the classification process; Region $UV$ [U+V+] consists of those spatially co-located pixels that have some role in both classification and adversarial attack; Region $\bar{U}V$ [U-V+] consists of those co-located pixels that are vulnerable towards an adversarial attack and are usually modified for the generation of an adversarial sample; Region $\bar{U}\bar{V}$ [U-V-] is the area in the image that is not important for either of the classification process or the generation of an adversarial sample. The mapping of each of the regions onto the plot of Utility and Vulnerability is presented in Figure \ref{split}. 


\subsection{Implications of isolating segments}
There are some key takeaways from the framework described above. Noting them will assist us in progressing towards designing the experimental setup necessary to validate this theoretical formulation. 

\subsubsection{Properties of segmentation}
Identifying $X_p$ and $X_q$, and thereby generating the RoI and the RoA respectively using the aforementioned manner is possible for any given image and its corresponding adversarial sample. While the \textit{Segregation} is universally possible for any given image and its corresponding adversarial sample, the spread of the RoI and the RoA on the image is subject to the choice of the thresholds $(\delta_1, \delta_2)$ and more generally on the distribution of the pixels in $X_p$ and $X_q$. This has to be verified empirically on real life datasets to check for the co-location property, which will offer evidence of the existence of well-defined regions. The study should be carried out on individual images and tested for if the observed patterns are consistent across multiple images of the class. A direct outcome of the above is the positioning of the isolated regions $UV, U\bar{V},  \bar{U}V ,\bar{U}\bar{V} $. A more bounded and well knit RoI and RoA will result in better segments, each being distinctive in nature. 


\subsubsection{Neutralizing region $\bar{U}V$ [U-V+]}
As evident from the formulation, the region $\bar{U}V$ is particularly vulnerable towards adversarial attacks. It is highly non-robust and vulnerable and also not useful for the task of classification. It is intuitive to therefore make use of this region and somehow block it from being picked up by the classifier in order to prevent adversarial mis-classification. This idea led us to develop a post-hoc adversarial defense mechanism. In theory, eliminating the information content in the region $\bar{U}V$, which we call the \textit{Neutralization} process should be effective in prohibiting further adversarial vulnerability. One may choose to extract a blob, comprising of the overlapping areas of the region $\bar{U}V$ of all images in a particular class, and use it on potential test samples to avert mis-classification. We have tested this idea thoroughly in Section \ref{experiments}.

Furthermore, it may be mentioned that the choice of $\delta_1$ and $\delta_2$ for generating the RoI and the RoA should maximize the classification accuracy of the trained classifier and also maximally weakens the potential adversarial attack. This can thought of as a tuning parameter for each dataset. 

\section{Investigating Co-located Spatial Patterns} \label{patterns}
\begin{figure}[ht]
\centering
\includegraphics[width=0.485\textwidth]{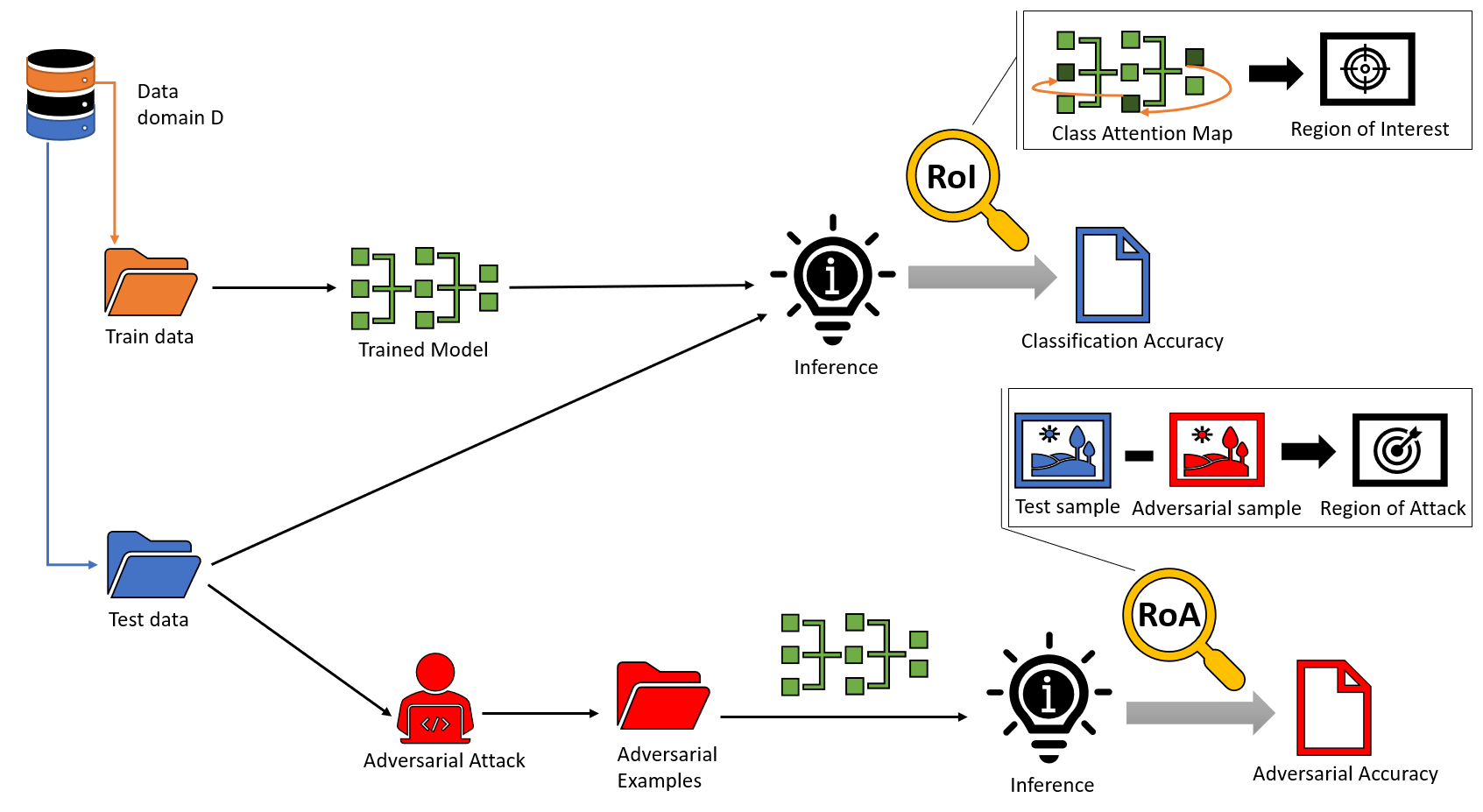}
\caption{\textit{Segregation}: Generating the RoI and the RoA for the task of classification and adversarial attack respectively. }
\label{segregation}
\end{figure}

In order to accomplish the objective of identifying and explaining the spatial patterns in the input space, that affect adversarial attacks, we have developed a pipeline that helps us test our assertions. The pipeline consists of three stages: \textit{Segregation}, \textit{Isolation} and \textit{Neutralization}.  We consider a binary classification problem (which has a fixed pathway of adversarial examples generation), its corresponding trained model $M$ and an adversarial attack mechanism $A$, which has a set $\epsilon$ as its hyper-parameter. This remains the same throughout. The first component of the pipeline is to map the features from the feature space to the input space, that is the pixels of the image. To identify the RoI at the individual level for each image, we use the technique of attention maps and identify the important features necessary for the classification task and project them back onto the input space. To identify the RoA at the individual level for each image, we look at the pixels that are modified by the attack module to generate the adversarial sample. This process is graphically represented in Figure \ref{segregation}. 

\begin{figure}[htbp]
\centering
\includegraphics[width=0.485\textwidth]{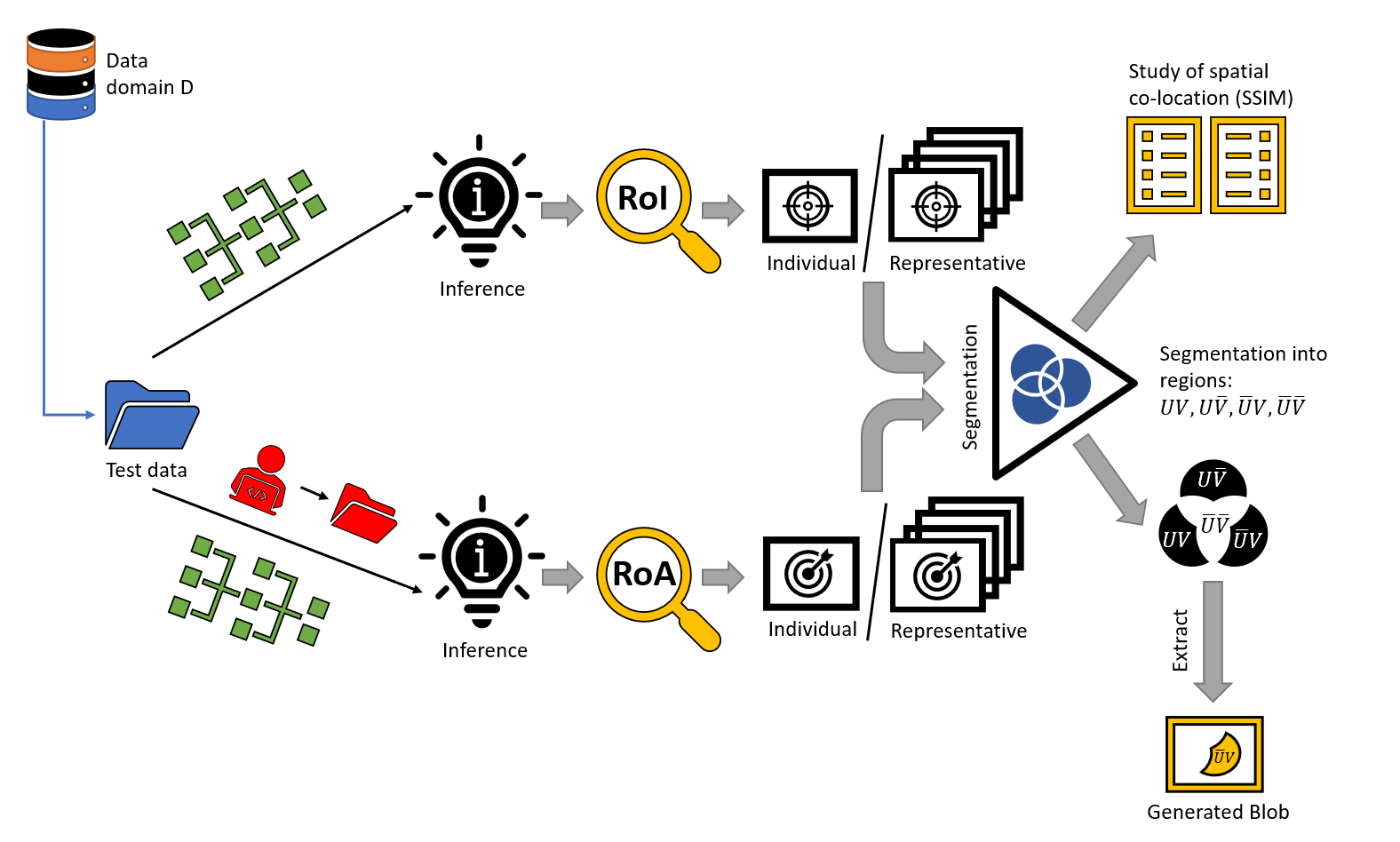}
\caption{\textit{Isolation} of regions $UV, U\bar{V},  \bar{U}V , \bar{U}\bar{V} $}
\label{isolation}
\end{figure}

The representative RoI is formed by considering the union of all pixels contributed by the individual images, meaning a superimposition or stacking up of the individual ROI for them. The same approach is also used to generate the representative RoA. 
Upon the identification of the RoI and the RoA, the second part of the pipeline carries out the \textit{Isolation} of the segments as mentioned in the earlier section and pictorially described in Figure \ref{isolation}.

\begin{figure}[htbp]
\centering
\includegraphics[width=0.485\textwidth]{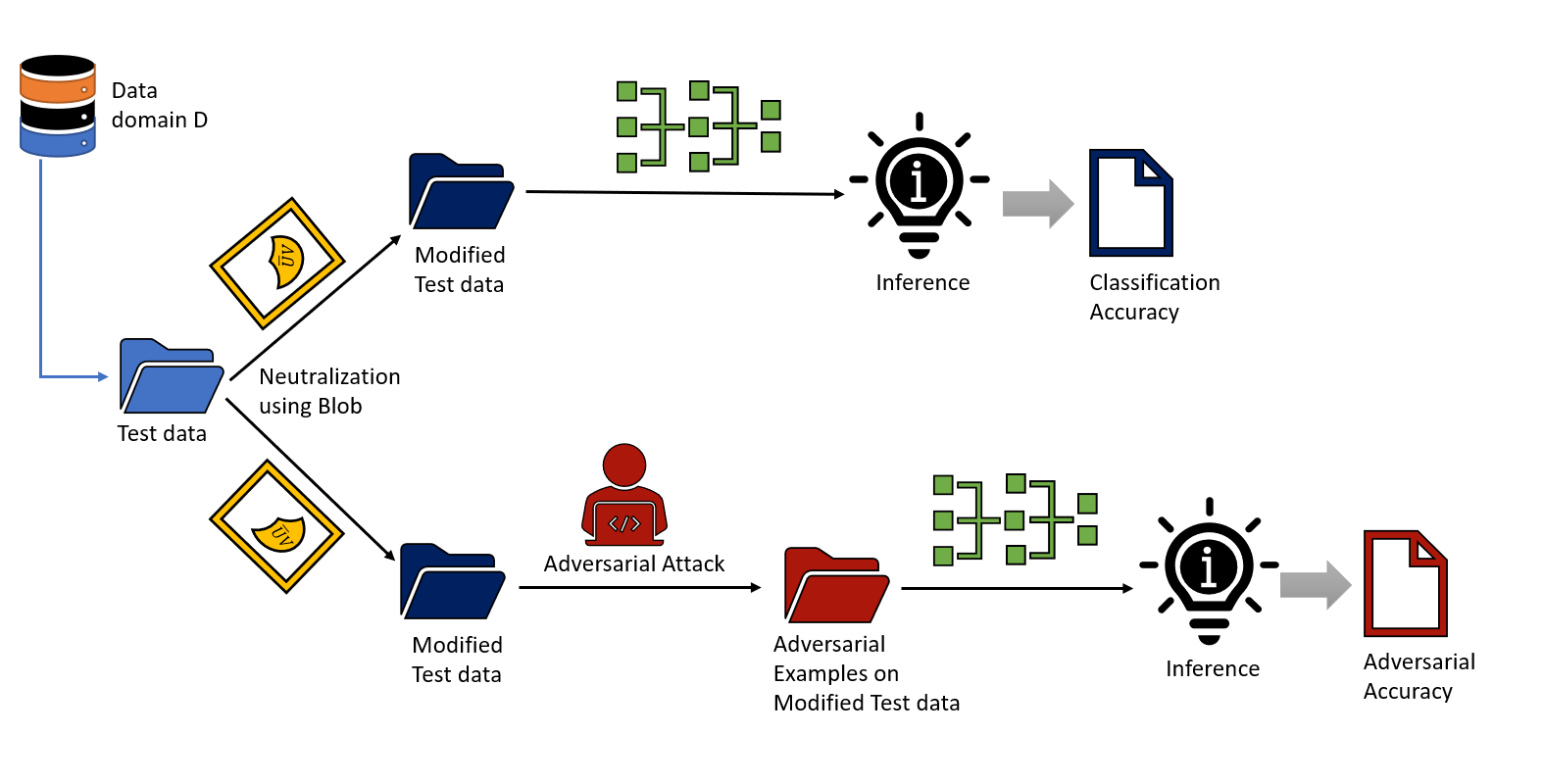}
\caption{Effect of \textit{Neutralization} on classification and further adversarial vulnerability}
\label{defense}
\end{figure}

 The third part of the pipeline is designed to study the post-hoc adversarial defense mechanism. The structure of analysis is as follows:  We begin with the clean set of test samples $S_{clean}$ and note the accuracy of $M$ on it, and refer to it as $score_{clean}$. Then we use attack module $A$ on $S_{clean}$ to generate adversarial examples $S_{adv}$, and check the performance of $M$ on it, referred to as $score_{adv}$. Post an adversarial attack, naturally we expect the accuracy $score_{adv}$ to be pretty low.  We perform the \textit{segregation} operation by generating the RoI and the RoA on $S_{clean}$, by making use of both $S_{clean}$ and $S_{adv}$. Thereafter, we obtain the regions $UV, U\bar{V},  \bar{U}V , \bar{U}\bar{V} $. At the class representative level we set a threshold $\theta$ for the determination of the region $\bar{U}V$ and extract the blob out. We note its spatial location (position of pixel indices) and use it to create $S_{mod}$, which is the modified version of $S_{clean}$, wherein the particular pixels in the location of the blob have been set to $0$ or $1$, to eliminate any variation (information) therein, therefore \textit{neutralizing} those pixels. We use the original model $M$ on $S_{mod}$ and note its performance $score_{mod}$ to see what impact the \textit{Neutralization} process has on classification accuracy. For the final part, we make use of the adversarial attack module $A$ with same $\epsilon$ on $S_{mod}$ to generate further adversarial examples $S_{mod-adv}$, as shown in Figure \ref{defense}. We use the model $M$ on $S_{mod-adv}$ to see its performance, referred to as $score_{mod-adv}$. Intuitively, if our proposed post-hoc adversarial defense mechanism works, then we expect $score_{clean}$ and $score_{mod}$ to be comparable, meaning the addition of blob doesn't impact classification much as it alters region $\bar{U}V$ primarily, whereas $score_{mod-adv}$ to be greater than $score_{adv}$ meaning the adversarial attack on the neutralized samples has become weaker. 

The key aspects of investigation through the utilization of the aforementioned pipeline are listed below. Obeying the key intuition of specificity between the paths of adversarial attack among trained manifolds of particular classes, namely a `Source' class and a `Target' class, we study the following.  
\begin{itemize}
    \item the co-location of pixels that contribute towards classification, therefore the existence of the Region of Interest (RoI) for individual images
    \item the co-location of pixels that contribute towards the adversarial attack, therefore the existence of the Region of Attack (RoA) for individual images
    \item the co-location of pixels that contribute towards classification when images of a particular class are aggregated (superimposed), therefore the existence of the Region of Interest (RoI) for the representative at the class level
    \item the co-location of pixels that contribute towards adversarial attack when images of a particular class are aggregated (superimposed), therefore the existence of the Region of Attack (RoA) for the representative at the class level
    \item similarity in spatial positioning (overlap) of the RoI and the RoA for each individual image and for the representatives at the class level
    \item segregating the region $\bar{U}V$ [U-V+] at the class representative level to generate the blob (to be used for \textit{Neutralization}) by choosing a threshold of overlap
    \item effect of \textit{Neutralization} on classification accuracy, that is using the generated blob for the class of images to neutralize the images at the individual level, and thereafter passing through the trained classifier
    \item effect of \textit{Neutralization} on further adversarial vulnerability, that is using the generated blob for the class of images to neutralize the images at the individual level, and thereafter carrying out further adversarial attack on the modified images and check their performance
\end{itemize}

\section{Experimental Results} \label{experiments}
In this section, we present the summary of our results of extensive experimentation to verify the propositions made in the earlier sections. In general, we used standard datasets and a well known image classifier VGG16 ~\cite{vgg} for ensuring that the results are comparable and reproducible. The datasets used include MNIST ~\cite{mnist}, Fashion-MNIST ~\cite{fashion}, Cifar-10 ~\cite{cifar}, the Cats vs Dogs dataset ~\cite{catsdogs} and the Malaria dataset ~\cite{malaria}. For the generation of the adversarial attacks, we used the well documented FGSM ~\cite{FSGM} attack module consistently throughout our experiments.

\subsection{Observations} \label{obs}
Due to constraints of content, we have not been able to report all results within the paper and they are made available for perusal at the GitHub repository \href{https://github.com/frusdelion/demystifying-adversarial-attacks}{here}. For understanding and putting forward our argument, we present a subset of the results. We consider two classes each for MNIST (0 and 1) and Fashion-MNIST (Trousers and Pullovers) randomly, and study it as a binary classification problem. The rest two are binary classification problems by default. Therefore, the snapshot of results presented hereafter are for the following: (a) MNIST - two classes (0 and 1), (b) Fashion MNIST - two classes (Trousers and Pullovers), (c) Cats vs Dogs - two classes (Cats and Dogs) and (d) Malaria - two classes (Parasitized and Uninfected).

\subsubsection{To study the existence of a co-located Region of Interest (RoI) for individual images}  
\begin{figure}[!ht]
\centering
\includegraphics[width=0.48\textwidth]{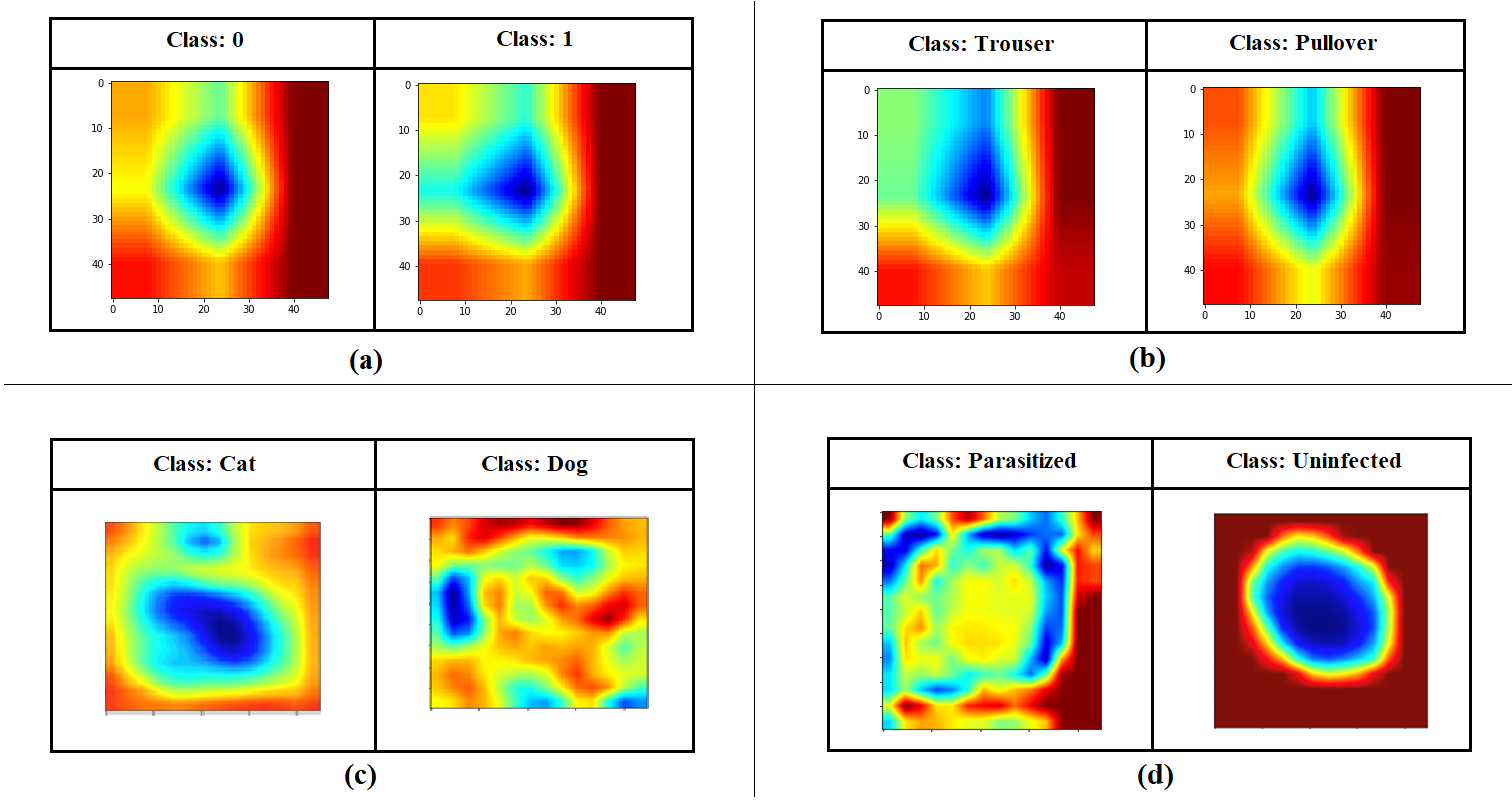}
\caption{Study of Existence of RoI in individual images for the datasets: (a) MNIST, (b) Fashion MNIST, (c) Cats vs Dogs and (d) Malaria. }
\label{ROI_EX_INDV}
\end{figure}

We set a tuned parameter for threshold $\delta_1$ and generate the RoI for all images in the class after classifying the images using the trained model $M$ and using the class attention map. We carry out a physical overview of it and for documentation, select one image at random and present its RoI. This process is then repeated for all classes in the dataset. The sample results are presented in Figure \ref{ROI_EX_INDV}. For each of the examples, the region marked in blue is the region corresponding to area of highest leverage on the classifier for the task of classification. 

\subsubsection{To study the existence of a co-located Region of Attack (RoA) for individual images} 
\begin{figure}[!ht]
\centering
\includegraphics[width=0.48\textwidth]{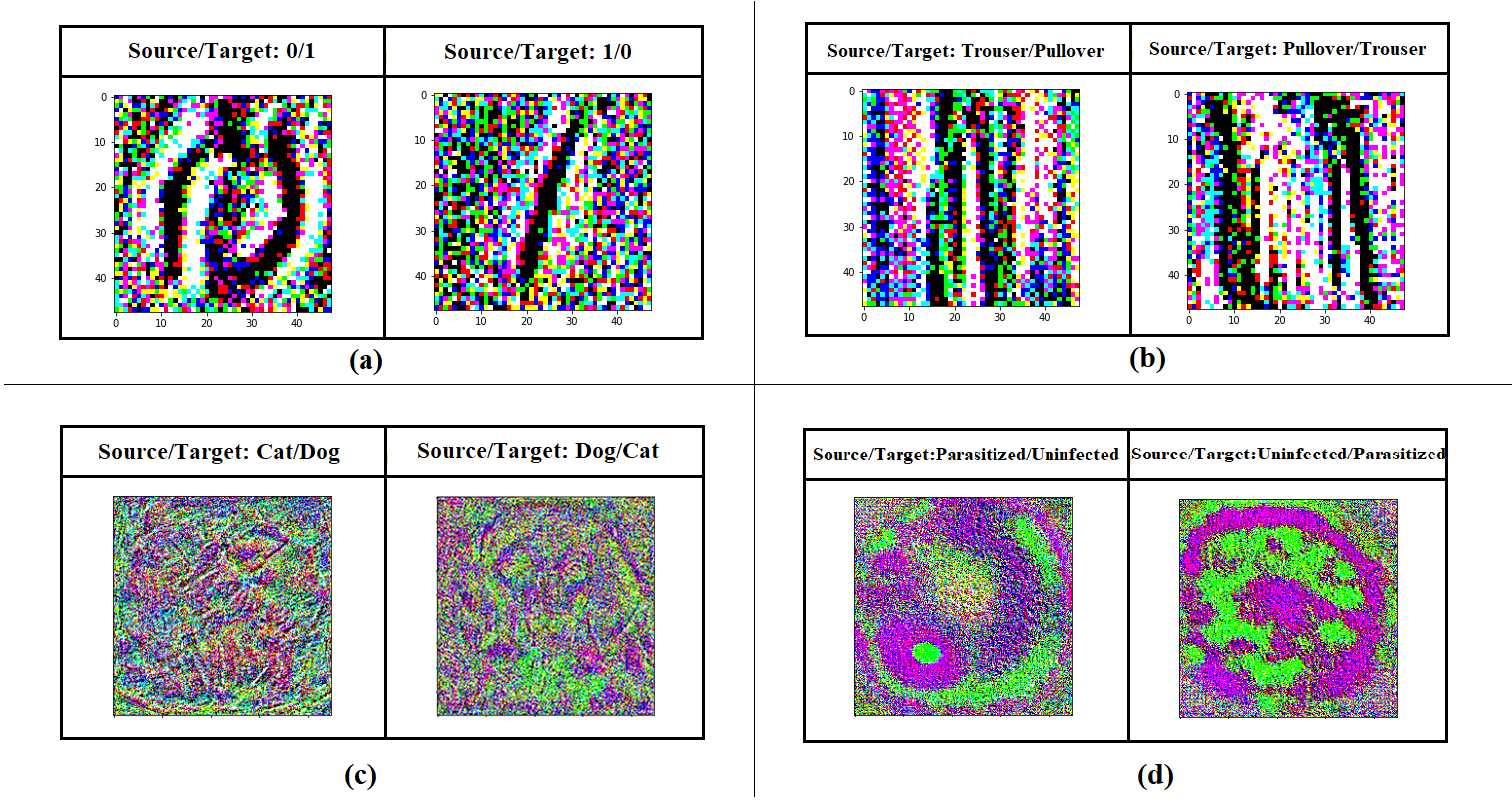}
\caption{Study of Existence of RoA in individual images for the datasets: (a) MNIST, (b) Fashion MNIST, (c) Cats vs Dogs and (d) Malaria.}
\label{ROA_EX_INDV}
\end{figure}

We set a tuned parameter for threshold $\delta_2$ and generate the RoA for all images in the class after carrying out an adversarial attack on it using an attack module $A$. We carry out a physical overview of it and for documentation, select one image at random and present its RoA. This process is then repeated for all classes in the dataset. The sample results are presented in Figure \ref{ROA_EX_INDV}. For (a) MNIST and (b) Fashion MNIST, the regions in black are corresponding to the regions maximally modified by the adversarial perturbation. For (c) Cats vs Dogs and (d) Malaria, the region of RoA are in violet. 

\subsubsection{To study the existence of a co-located Region of Interest (RoI) for all images belonging to a particular class}
\begin{figure}[!ht]
\centering
\includegraphics[width=0.48\textwidth]{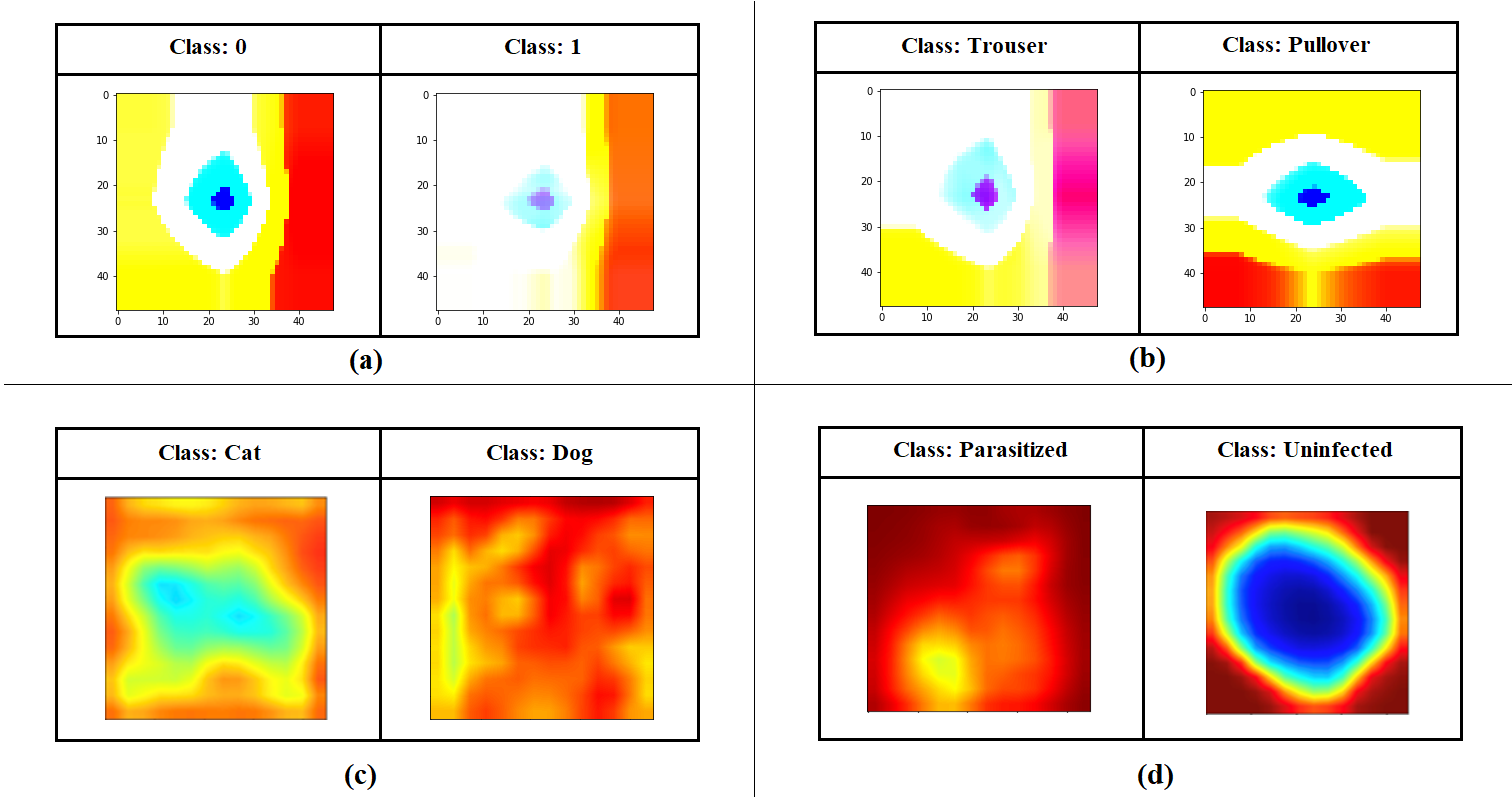}
\caption{Study of Existence of RoI at the class representative level for the datasets: (a) MNIST, (b) Fashion MNIST, (c) Cats vs Dogs and (d) Malaria.}
\label{RoI_EX_REP}
\end{figure}

After the generation of the individual RoI, the images are superimposed and the mean value is considered pixel-wise to create the representative RoI for the class, which is then presented in Figure \ref{RoI_EX_REP}. This is to look for the existence of overlap between the RoI of images in the class. The colour code naturally remains the same as that of the individually identified RoI.

\subsubsection{To study the existence of a co-located Region of Attack (RoA) for all images belonging to a particular class}
\begin{figure}[!ht]
\centering
\includegraphics[width=0.48\textwidth]{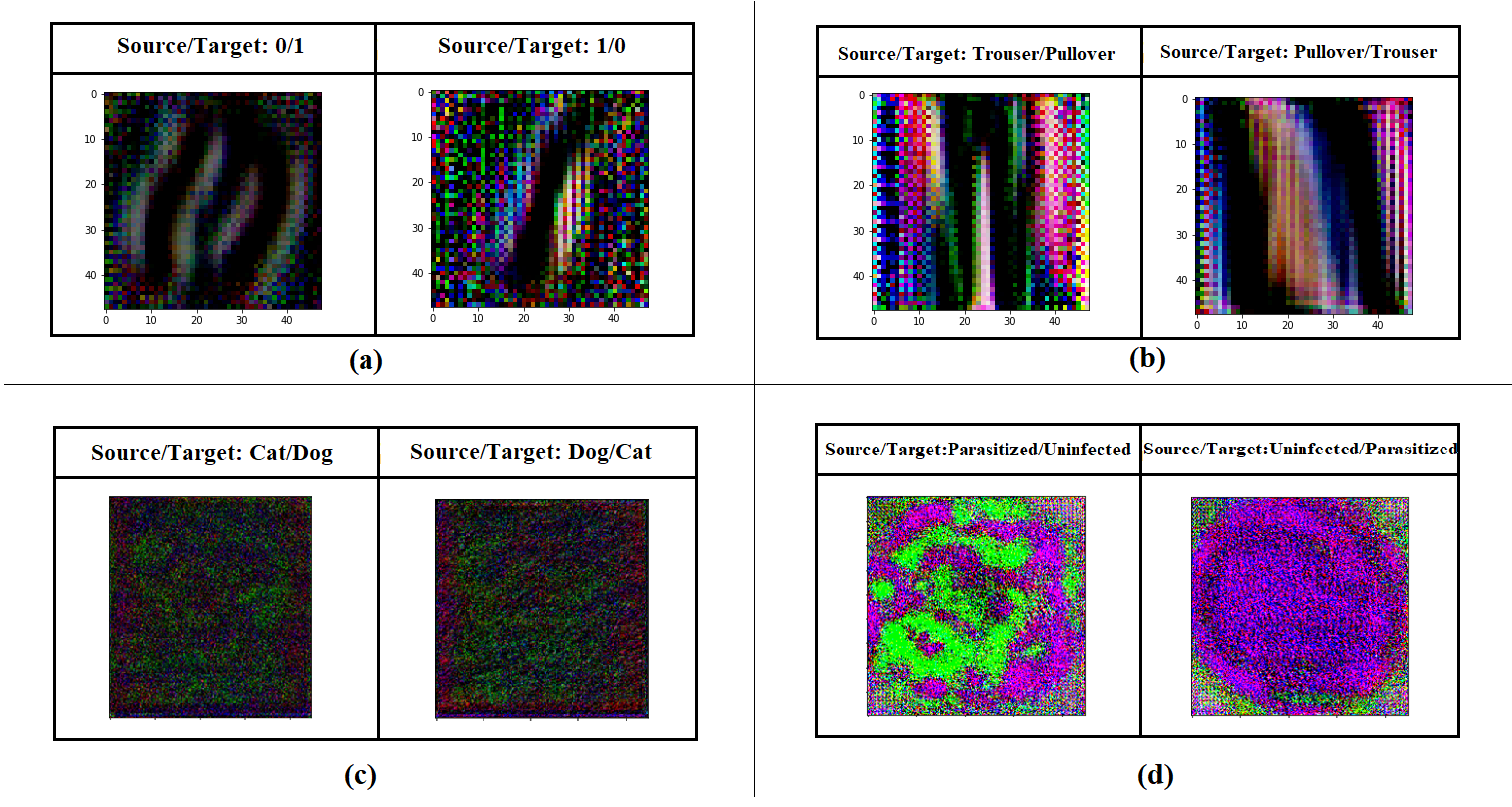}
\caption{Study of Existence of RoA at the class representative level for the datasets: (a) MNIST, (b) Fashion MNIST, (c) Cats vs Dogs and (d) Malaria .}
\label{RoA_EX_REP}
\end{figure}

After the generation of the individual RoA, the images are superimposed and the mean value is considered pixel-wise to create the representative RoA for the class, which is then presented in Figure \ref{RoA_EX_REP}. This is to look for the existence of overlap between the RoA of images within the class. 

\subsubsection{To study the overlap in spatial positioning of the RoI and RoA for individual images and for the representatives at the class level} 
For every image belonging to the particular class, the RoI and the RoA are already determined. To study their spatial positioning and overlap, the pairwise structural similarity is measured using SSIM and class-level mean is reported. For the representatives at the class level, the SSIM of the corresponding representative versions is reported in Table \ref{ssim}. 

\begin{table}[!ht]
\caption{Structural similarity of RoI and RoA}
\label{ssim}
\begin{tabular}{c|c|c|c}
\hline
\multirow{2}{*}{Dataset} &
  \multirow{2}{*}{\begin{tabular}[c]{@{}c@{}}Source Class /\\ Target Class\end{tabular}} &
  \multirow{2}{*}{\begin{tabular}[c]{@{}c@{}}SSIM value\\ (mean of samples)\end{tabular}} &
  \multirow{2}{*}{\begin{tabular}[c]{@{}c@{}}SSIM value \\ (representatives)\end{tabular}} \\
                                                                          &                      &        &         \\ \hline \hline 
\multirow{2}{*}{MNIST}                                                    & 0 to 1               & 0.2178 & 0.0039  \\ \cline{2-4} 
                                                                          & 1 to 0               & 0.1978 & 0.0001  \\ \hline
\multirow{2}{*}{\begin{tabular}[c]{@{}c@{}}Fashion \\ MNIST\end{tabular}} & Trousers to Pullover & 0.1923 & -0.0019 \\ \cline{2-4} 
                                                                          & Pullover to Trousers & 0.2029 & -0.0003 \\ \hline
\multirow{2}{*}{\begin{tabular}[c]{@{}c@{}}Cats vs \\ Dogs\end{tabular}}  & Cats to Dogs         & 0.3028 & 0.0899  \\ \cline{2-4} 
                                                                          & Dogs to Cats         & 0.3046 & 0.0938  \\ \hline
\end{tabular}
\end{table}

\subsubsection{Generation of neutralizer blob} 
\begin{figure}[!ht]
\centering
\includegraphics[width=0.48\textwidth]{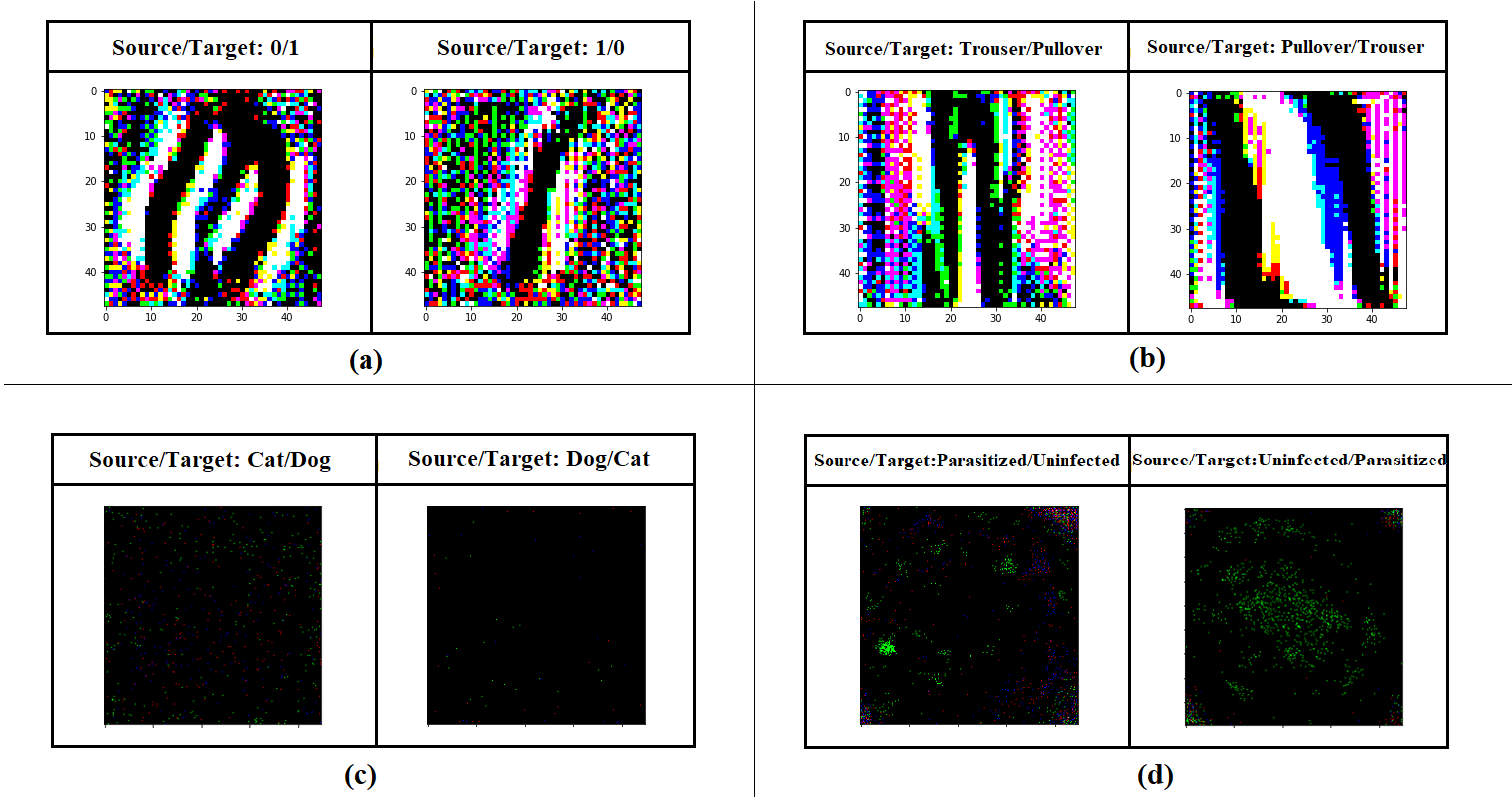}
\caption{Blobs for the datasets: (a) MNIST, (b) Fashion MNIST, (c) Cats vs Dogs  and (d) Malaria.}
\label{blob}
\end{figure}

To demonstrate the process of \textit{Neutralization}, we \textit{isolate} the region $\bar{U}V$ [U-V+] for each class of images. This generates the blob that indicates the region of highest adversarial vulnerability. The \textit{Neutralization} process sets all the pixels to its highest/lowest possible value, without a loss of generality, to eliminate its specific information content. The generated blobs are presented in Figure \ref{blob}. 
     
\subsubsection{To study the effect of \textit{Neutralization} on classification accuracy}
We report $score_{clean}$ and $score_{mod}$ for the four datasets in Table \ref{clean-mod}.
   
\begin{table}[!ht]
\caption{Effect of \textit{Neutralization} (blob based, \textit{isolated} vulnerable region) on the task of classification.  }
\label{clean-mod}
\begin{tabular}{c|c|c|c}
\hline
\multirow{2}{*}{Dataset} & \multirow{2}{*}{\begin{tabular}[c]{@{}c@{}}Source Class /\\ Target Class\end{tabular}} & \multicolumn{2}{c}{Classification accuracy ($M$)} \\ \cline{3-4} 
                                                                         &                           & $score_{clean}$ & $score_{mod}$ \\ \hline \hline 
\multirow{2}{*}{MNIST}                                                   & 0 to 1                    & 100\%           & 93.75\%       \\ \cline{2-4} 
                                                                         & 1 to 0                    & 100\%           & 96.88\%       \\ \hline
\multirow{2}{*}{\begin{tabular}[c]{@{}c@{}}Fashion\\ MNIST\end{tabular}} & Trousers to Pullover      & 100\%           & 96.88\%       \\ \cline{2-4} 
                                                                         & Pullover to Trousers      & 100\%           & 96.88\%       \\ \hline
\multirow{2}{*}{\begin{tabular}[c]{@{}c@{}}Cats vs\\ Dogs\end{tabular}}  & Cats to Dogs              & 98.97\%         & 85.35\%       \\ \cline{2-4} 
                                                                         & Dogs to Cats              & 99.87\%         & 99.89\%       \\ \hline
\multirow{2}{*}{Malaria}                                                 & Parasitized to Uninfected & 99.79\%         & 99.68\%       \\ \cline{2-4} 
                                                                         & Uninfected to Parasitized & 99.21\%         & 99.56\%       \\ \hline
\end{tabular}
\end{table}

\subsubsection{To study the effect of \textit{Neutralization} on further adversarial vulnerability}
We report $score_{adv}$ and $score_{mod-adv}$, for the four datasets in Table \ref{mod-adv}.

\begin{table}[!ht]
\caption{Effect of \textit{Neutralization} (blob based, for vulnerable region) on the potential of further adversarial attack.  }
\label{mod-adv}
\begin{tabular}{c|c|c|c}
\hline
\multirow{2}{*}{Dataset} &
  \multirow{2}{*}{\begin{tabular}[c]{@{}c@{}}Source Class /\\ Target Class\end{tabular}} &
  \multicolumn{2}{c}{Classification accuracy ($M$)} \\ \cline{3-4} 
                         &                           & $score_{adv}$ & $score_{mod-adv}$ \\ \hline \hline 
\multirow{2}{*}{MNIST}   & 0 to 1                    & 90.62\%       & 100\%             \\ \cline{2-4} 
                         & 1 to 0                    & 96.85\%       & 100\%             \\ \hline
\multirow{2}{*}{\begin{tabular}[c]{@{}c@{}}Fashion \\ MNIST\end{tabular}} &
  Trousers to Pullover &
  90.62\% &
  90.62\% \\ \cline{2-4} 
                         & Pullover to Trousers      & 96.85\%       & 100\%             \\ \hline
\multirow{2}{*}{\begin{tabular}[c]{@{}c@{}}Cats vs \\ Dogs\end{tabular}} &
  Cats to Dogs &
  3.47\% &
  3.49\% \\ \cline{2-4} 
                         & Dogs to Cats              & 7.37\%        & 2.67\%            \\ \hline
\multirow{2}{*}{Malaria} & Parasitized to Uninfected & 5.85\%        & 46.11\%           \\ \cline{2-4} 
                         & Uninfected to Parasitized & 7.55\%        & 45.06\%           \\ \hline
\end{tabular}
\end{table}

\subsection{Key Findings} \label{findings}
The key aspects to note, based on the overall experimental results, are as follows:
\begin{itemize}
    \item From the study of the existence of the Region of Importance (RoI) and the Region of Attack (RoA), both at the individual level and at the class representative level, we understand that the segregation is possible, as there exists a good co-location of pixels, which form observable clusters, with specific significance for each as explained. It is more apparent than others in some cases, depending on the particular datasets and their inherent complexities. It must be noted that the co-location of regions at the representative level holds good only for samples which are free of translational variance. 
    \item Upon the identification of the the regions, the splitting up of the image's input space into four disjoint segments ($UV, U\bar{V},  \bar{U}V , \bar{U}\bar{V} $) is also feasible. The shapes and sizes of these parts, vary from dataset to dataset, and is highly dependant on the overlap that exists between the RoI and RoA for that particular class of images.
    \item Extending the findings to adopt a practical use of these patterns, the idea of attempting to neutralize the thus-identified vulnerable areas (position demarcated by the blob), could be done in multiple ways. One way, as adopted in this work, is to eliminate the information content of those pixels by \textit{Neutralization}, thereby reducing vulnerability. 
    \item The proposed post-hoc defense mechanism, obtained by using the segmentation of regions, works. The experimental results show that while the act of \textit{Neutralization} does not affect the task of classification much, it is able to reduce further vulnerabilities of the samples. This mechanism is more effective in some datasets than others. 
\end{itemize}

\section{Conclusions} \label{conclusion}
To study the patterns within the distribution of pixels within an image and how they may relate to the context of its classification and adversarial attacks, we found empirical evidence of spatial co-location among pixels that have particular significance. We have defined an Region of Importance (RoI) and an Region of Attack (RoA) that correspond to the areas in the image specifically responsible for the task of classification and rendering adversarial vulnerability. We have demonstrated through multiple datasets that although their positions and distribution vary, and their existence may be more apparent in some cases than the rest, these patterns can be used for a better understanding of specific adversarial attacks. In fact, this knowledge has been used to develop a post-hoc adversarial defense mechanism that tries to neutralize the regions of the image particularly vulnerable towards an attack. This is significant towards the reliable use of ML models. In the future, better modes of identifying the regions RoI and RoA followed by isolation may help us in improving the segregation which will benefit the neutralization born out of it. 

\bibliographystyle{IEEEtran}
\bibliography{reference}

\end{document}